\documentclass[]{interact}

\usepackage{epstopdf}
\usepackage[caption=false]{subfig}

\usepackage{natbib}
\bibpunct[, ]{(}{)}{;}{a}{}{,}

\usepackage{url}
\theoremstyle{plain}

\theoremstyle{definition}

\theoremstyle{remark}

\begin{document}

\articletype{ARTICLE}

\title{KST-Mixer: Kinematic Spatio-Temporal Data Mixer For Colon Shape Estimation}

\author{
\name{Masahiro Oda\textsuperscript{a,b}\thanks{CONTACT Masahiro Oda. Email: moda@mori.m.is.nagoya-u.ac.jp}
and Kazuhiro Furukawa\textsuperscript{c} and Nassir Navab\textsuperscript{d} and Kensaku Mori\textsuperscript{b,e}}
\affil{
\textsuperscript{a}Information and Communications, Nagoya University, Nagoya, Japan; 
\textsuperscript{b}Graduate School of Informatics, Nagoya University, Nagoya, Japan;
\textsuperscript{c}Department of Gastroenterology and Hepatology, Nagoya University Graduate School of Medicine, Nagoya, Japan;
\textsuperscript{d}Technical University of Munich, M\"{u}nchen, Germany;
\textsuperscript{e}Research Center for Medical Bigdata, National Institute of Informatics, Tokyo, Japan}
}

\maketitle

\begin{abstract}
We propose a spatio-temporal mixing kinematic data estimation method to estimate the shape of the colon with deformations caused by colonoscope insertion.
Endoscope tracking or a navigation system that navigates physicians to target positions is needed to reduce such complications as organ perforations.
Although many previous methods focused to track bronchoscopes and surgical endoscopes, few number of colonoscope tracking methods were proposed.
This is because the colon largely deforms during colonoscope insertion.
The deformation causes significant tracking errors.
Colon deformation should be taken into account in the tracking process.
We propose a colon shape estimation method using a Kinematic Spatio-Temporal data Mixer (KST-Mixer) that can be used during colonoscope insertions to the colon.
Kinematic data of a colonoscope and the colon, including positions and directions of their centerlines, are obtained using electromagnetic and depth sensors.
The proposed method separates the data into sub-groups along the spatial and temporal axes.
The KST-Mixer extracts kinematic features and mix them along the spatial and temporal axes multiple times.
We evaluated colon shape estimation accuracies in phantom studies.
The proposed method achieved 11.92 mm mean Euclidean distance error, the smallest of the previous methods.
Statistical analysis indicated that the proposed method significantly reduced the error compared to the previous methods.
\footnote[1]{Code and data of the proposed method are available at: \url{https://github.com/modafone/kst-mixer}}
\end{abstract}

\begin{keywords}
Colon; Colonoscope tracking; Kinematic data estimation; Multi layer perceptron
\end{keywords}

\section{Introduction}

CT colonography (CTC) is performed to find colonic polyps from CT images.
If colonic polyps or early-stage cancers are found in a CTC, a colonoscopic examination is performed to endoscopically remove them.
A physician controls the colonoscope based on its camera view during a colonoscope examination.
However, its viewing field is limited and unclear because the camera is often covered by fluid or the colonic wall.
Furthermore, the colon changes shape significantly during colonoscope insertion.
Physicians require much skill and experience to estimate how the colonoscope travels inside the colon.
Inexperienced physicians overlook polyps or cause such complications as colon perforation.
A colonoscope navigation system is needed that guides the physician to the polyp position.
A colonoscope tracking method is necessary as its core.

Many tracking methods of endoscopes are proposed \citep{peters08,deligianni05,rai08,deguchi09,visentini17,wang21,banach21,gildea06,schwarz06,mori05,luo15,liu13,yao21,ching10,fukuzawa15,oda17}.
They can be classified into the image-based, the sensor-based, and the hybrid methods.
The bronchoscope is the main application of bronchoscope tracking.
Many researchers proposed image- and sensor-based methods in recent several decades.
Image-based tracking methods estimate the camera movements based on image registrations.
Registrations between temporally continuous bronchoscopic images \citep{peters08} or between real and virtualized bronchoscopic images \citep{deligianni05,rai08,deguchi09} are used for tracking.
Recent approaches use deep learning-based depth estimation results to improve image-based tracking accuracy \citep{visentini17,wang21,banach21}.
Sensor-based tracking methods use small sensors to obtain bronchoscope position \citep{gildea06,schwarz06}.
Hybrid methods of them use both image and sensor information to accurately estimate bronchoscope position \citep{mori05,luo15}.
Some research groups propose tracking methods for colonoscope.
In colonoscope tracking, the image-based method \citep{liu13} is difficult to apply because unclear colonoscopic images appear frequently.
Unclear image removal and disparity map estimation are utilized to improve tracking accuracy \citep{yao21}.
Electromagnetic (EM) sensors are used to obtain colonoscope shapes \citep{ching10,fukuzawa15}.
Unfortunately, they cannot guide physicians to polyp positions because they cannot map the colonoscope shape to a colon in a CT volume, which may contain polyp detection results.
Combining the colonoscope position or shape information with polyp position information, which can be detected in a CT volume taken prior to colonoscope insertion, is essential in colonoscope navigation.
Such navigation is called colonoscope-CT-based navigation.

A few colonoscope tracking methods are applicable to perform colonoscope-CT-based navigation.
Reference \citep{oda17} used colonoscope shape measured by an EM sensor and CT volume for colonoscope tracking.
The method obtains two curved lines representing the colon and colonoscope shapes to estimate the colonoscope position on a CT volume coordinate system.
The method enables real-time tracking regardless of the colonoscopic image quality.
However, the method does not consider colon deformation caused by colonoscope insertion.
Such deformation caused significant tracking error.
Large tracking errors were observed at the transverse and sigmoid colons, which are significantly deformed by a colonoscope insertion.
To reduce tracking errors, estimation methods of the shape of the colon with deformations caused by colonoscope insertion were proposed \citep{oda18,oda18-2}.
One of them \citep{oda18} uses the shape estimation network that has a long short-term memory (LSTM) layer \citep{lstm} to estimate the colon shape and the other \citep{oda18-2} uses regression forests.
Estimation accuracies of them need to be improved to perform accurate colonoscope tracking.

We propose a novel shape estimation method of the colon for colonoscope tracking using a Kinematic Spatio-Temporal data Mixer (KST-Mixer).
The proposed method estimates the colon from time-series shape data of the colonoscope, which is inserted into the colon.
The KST-Mixer extracts kinematic features from the data of the colonoscope using simple multi-layer perceptrons (MLPs).
The extracted features are mixed along the spatial and temporal axes in spatio-temporal mixing blocks to estimate the colon shape dynamically.
Because the KST-Mixer has a simple processing flow, it provides estimation results in a short processing time.
It is suitable to be used in real-time colonoscope navigation systems.

Contributions of this paper are summarized as: {\bf (1)} to propose a novel colon shape estimation method that utilizes spatial and temporal kinematic features extraction and mixing processes, {\bf (2)} to enable short computation time in estimation used in real-time colonoscope tracking, and {\bf (3)} to achieve the smallest shape estimation error among the previous methods.

\section{Method}

\subsection{Overview}

The proposed method estimates the colon shape from the colonoscope shape.
They are time-series data measured at a specific time interval.
The KST-Mixer is trained to estimate a colon shape from time-series colonoscope shapes.
After the training, a trained model estimates a colon shape during a colonoscope insertion.

\subsection{Colon and colonoscope shape representation}

\begin{figure}[tb]
  \begin{minipage}{0.5\textwidth}
    \centering
\subfloat[]{\includegraphics[width=0.35\textwidth]{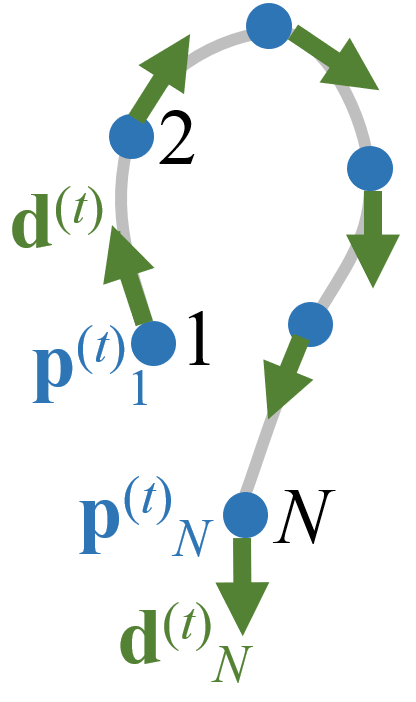}\label{fig:shape1}}
  \end{minipage}
  \begin{minipage}{0.5\textwidth}
\centering
\subfloat[]{\includegraphics[width=0.4\textwidth]{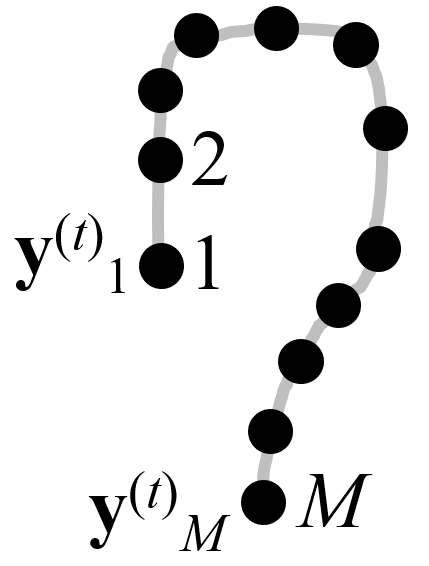}\label{fig:shape2}}
  \end{minipage}
\caption{Examples of (a) colonoscope and (b) colon shapes.} \label{fig:shape}
\end{figure}

We represent the colon and colonoscope shapes as point sets.
The colonoscope shape (Fig. \ref{fig:shape} (a)) is a set of 3D positions ${\bf p}^{(t)}_{n}$ and 3D directions ${\bf d}^{(t)}_{n}$, that is represented as
\begin{equation}
{\bf X}^{(t)} = \{ {\bf p}^{(t)}_{n}, {\bf d}^{(t)}_{n}; n=1, \ldots, N \},
\end{equation}
where $t \ (t=1, \ldots, T)$ is the index of time, $T$ is the total number of time frames, and $N$ is the number of points in the colonoscope shape.
${\bf p}^{(t)}_{n}$ is a point aligned along the colonoscope centerline.
${\bf p}^{(t)}_{1}$ corresponds to the position of the colonoscope tip.
${\bf d}^{(t)}_{n}$ is a tangent direction of the colonoscope tube at ${\bf p}^{(t)}_{n}$.

The colon shape (Fig. \ref{fig:shape} (b)) is a set of 3D points ${\bf y}^{(t)}_{m}$ that is represented as
\begin{equation}
{\bf Y}^{(t)} = \{ {\bf y}^{(t)}_{m}; m=1, \ldots, M \},
\end{equation}
where $M$ is the number of points in the colon shape.
${\bf y}^{(t)}_{m}$ is a point aligned along the colon centerline.
${\bf y}^{(t)}_{1}$ and ${\bf y}^{(t)}_{M}$ correspond to the cecum and the anus positions, respectively.

\subsection{Kinematic spatio-temporal data mixer (KST-Mixer)}

\subsubsection{Overview of KST-Mixer}
The KST-Mixer estimates a colon shape from time-series colonoscope shapes.
Its architecture is based on MLPs that are repeatedly applied across the spatial or temporal axes.
This architecture is inspired by the MLP-Mixer \citep{mlpmixer}, which classifies images utilizing spatial locations and image features.
The MLP-Mixer has competitive image classification performance to current methods such as Vision Transformers (ViT) \citep{vit} and provides a short processing time.
We utilize the MLP-based architecture to process kinematic data in shape estimation tasks.

\subsubsection{Data preparation}
To generate input data of the KST-Mixer, we rearrange a time-series colonoscope shape data as a matrix with spatial and temporal axes.
Components of the 3D point and direction are represented as ${\bf p}^{(t)}_{n} = (p^{(t)}_{n,x}, p^{(t)}_{n,y}, p^{(t)}_{n,z})$ and ${\bf d}^{(t)}_{n} = (d^{(t)}_{n,x}, d^{(t)}_{n,y}, d^{(t)}_{n,z})$.
From ${\bf p}^{(t)}_{n}$, we define the positional matrix ($3N \times \tau$) of time period $t=t_{c}, \ldots, t_{c}-\tau+1$ as
\begin{equation}
{\bf P}^{(t_{c})} = 
\begin{pmatrix}
p^{(t_{c})}_{1,x} & \cdots & p^{(t_{c}-\tau+1)}_{1,x} \\
p^{(t_{c})}_{1,y} & \cdots & p^{(t_{c}-\tau+1)}_{1,y} \\
p^{(t_{c})}_{1,z} & \cdots & p^{(t_{c}-\tau+1)}_{1,z} \\
\vdots & \ddots & \vdots \\
p^{(t_{c})}_{N,x} & \cdots & p^{(t_{c}-\tau+1)}_{N,x} \\
p^{(t_{c})}_{N,y} & \cdots & p^{(t_{c}-\tau+1)}_{N,y} \\
p^{(t_{c})}_{N,z} & \cdots & p^{(t_{c}-\tau+1)}_{N,z} \\
\end{pmatrix},
\end{equation}
where $\tau$ is time length.
We also define the directional matrix as ${\bf D}^{(t_{c})}$ similarly.

\begin{figure}[tb]
\includegraphics[width=0.98\textwidth]{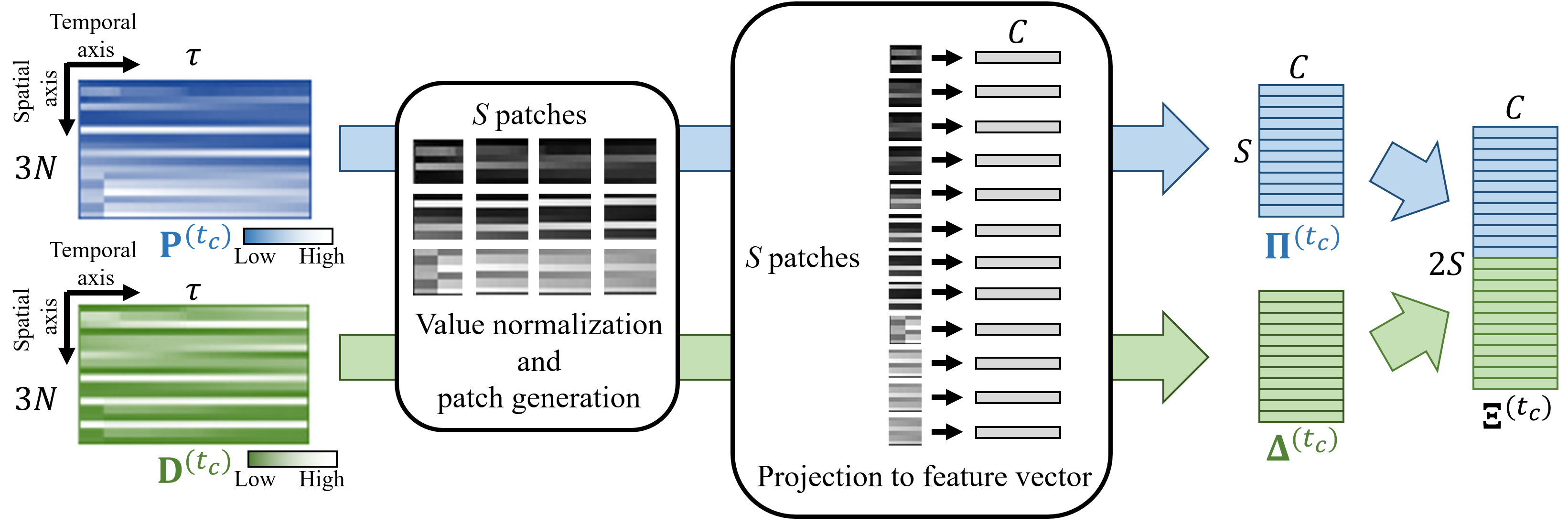}\label{fig:data}
\caption{Data preparation process from positional and directional matrixes ${\bf P}^{(t_{c})}, {\bf D}^{(t_{c})}$ to ${\bf \Xi}^{(t_{c})}$.} \label{fig:dataprep}
\end{figure}

Values in ${\bf P}^{(t_{c})}$ are normalized to take values in the range $[0, 1]$.
We regard the normalized matrix as a 2D image to generate $S$ non-overlapping and homogeneous sized image patches.
The image size is $(3N, \tau)$ and the size of each patch is $(s_{1}, s_{2})$.
From them, the number of patches is calculated as $S=\frac{3N \tau}{s_{1}s_{2}}$.
Each patch contains spatially and temporally local data.
Each patch is projected to a feature vector of hidden dimension $C$.
As the result, we obtain a input matrix of positional data ${\bf \Pi}^{(t_{c})} \in \mathbb{R}^{S \times C}$.
The order of feature values in the matrix is sensitive to both spatial and temporal axes.
Therefore, such a position embedding technique as ViT employs is not necessary.
We also obtain a input matrix of directional data ${\bf \Delta}^{(t_{c})}$ from ${\bf D}^{(t_{c})}$.
We make a matrix of colonoscope shape data ${\bf \Xi}^{(t_{c})} \in \mathbb{R}^{2S \times C}$ consisting of ${\bf \Pi}^{(t_{c})}$ and ${\bf \Delta}^{(t_{c})}$ elements.
This process is illustrated in Fig. \ref{fig:dataprep}.

We generate additional input data of the KST-Mixer from the insertion length of the colonoscope.
The insertion length of colonoscope at time $t$ is $l^{(t)}$.
A set of insertion lengths in the period of time $t=t_{c},\ldots,t_{c}-\tau+1$ is represented as ${\bf L}^{(t_{c})} = ( l^{(t_{c})} \cdots l^{(t_{c}-\tau+1)} )^{T}$, which is a column vector of size $\tau \times 1$.
The insertion length data ${\bf L}^{t_{c}}$ is used in the process of the KST-Mixer.

\subsubsection{Architecture of KST-Mixer}

\begin{figure}[tb]
\begin{center}
\includegraphics[width=0.98\textwidth]{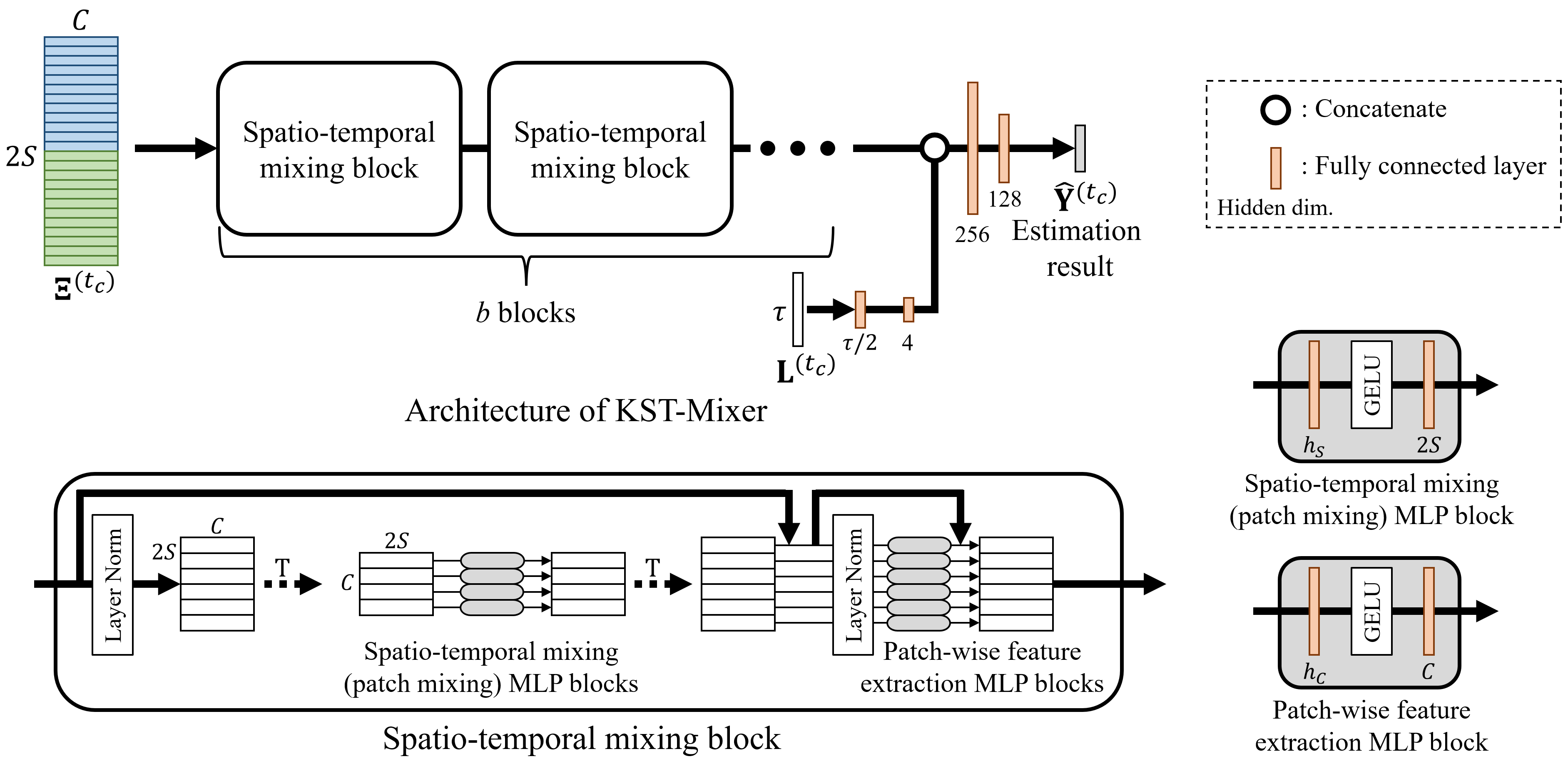}
\caption{Architecture of KST-Mixer. It estimates colon shape ${\bf \hat{Y}}^{t_{c}}$ from colonoscope shape data ${\bf \Xi}^{t_{c}}$ and insertion length data ${\bf L}^{t_{c}}$. It extracts colonoscope shape features in multiple spatio-temporal mixing blocks. Features are then combined with insertion length feature. Fully connected layers process combined features to generate estimation results.} \label{fig:architecture}
\end{center}
\end{figure}

The input of the KST-Mixer are ${\bf \Xi}^{(t_{c})}$ and ${\bf L}^{(t_{c})}$. The KST-Mixer outputs an estimated colon shape ${\bf \hat{Y}}^{(t_{c})}$.
The architecture of the KST-Mixer is shown in Fig. \ref{fig:architecture}.
It has $b$ spatio-temporal mixing blocks.
Each of them consists of two MLP blocks similarly to the MLP-Mixer \citep{mlpmixer}.
The first is the spatio-temporal feature mixing MLP block (patch mixing MLP block).
In the block, an input patch-wise feature vector is transposed and processed by the MLP block.
The second MLP block is the patch-wise feature extraction MLP block.

Each MLP block has two fully connected layers and an activation function.
Dropout with a probability of $d_{1}$ is performed.
Operation in the spatio-temporal mixing MLP block can be represented as
\begin{eqnarray}
\label{eq:mlp1} {\bf U}_{*,i} &=& {\bf I}_{*,i} + {\bf w}_{2} \sigma ( {\bf w}_{1} N({\bf I})_{*,i} ), \\
\label{eq:mlp2} {\bf O}_{j,*} &=& {\bf U}_{j,*} + {\bf w}_{4} \sigma ( {\bf w}_{3} N({\bf U})_{j,*} ),
\end{eqnarray}
where ${\bf I}, {\bf O}$ are input and output feature vectors, ${\bf w}_{1,\ldots,4}$ are weight parameters of fully connected layers, and $i=1,\ldots,C$, $j=1,\ldots,2S$.
$N$ is a layer normalization function \citep{layernorm}.
$\sigma$ is an GELU activation function \citep{gelu}.
$*$ indicates the row or the column vectors where operations are applied.
Equation (\ref{eq:mlp1}) is the calculation in the patch mixing MLP block. The calculation is performed for each column of ${\bf I}$.
The number of hidden units of the first fully connected layer in this block $h_{S}$ is used to control patch mixing.
Equation (\ref{eq:mlp2}) is the calculation in the patch-wise feature extraction MLP block. The calculation is performed for each rows of ${\bf U}$.
The number of the hidden units of the first fully connected layer in this block $h_{C}$ is used to control feature extraction from the patch.

After the processes of the spatio-temporal mixing blocks, feature values are mapped to a vector.
It is combined with feature values calculated from the insertion length data ${\bf L}^{(t_{c})}$ and then processed by some fully connected layers.
Dropout with a probability of $d_{2}$ is performed here.
The last layer outputs an estimated colon shape ${\bf \hat{Y}}^{(t_{c})}$.

\section{Experimental Setup}

\begin{figure}[tb]
\begin{center}
\includegraphics[width=0.98\textwidth]{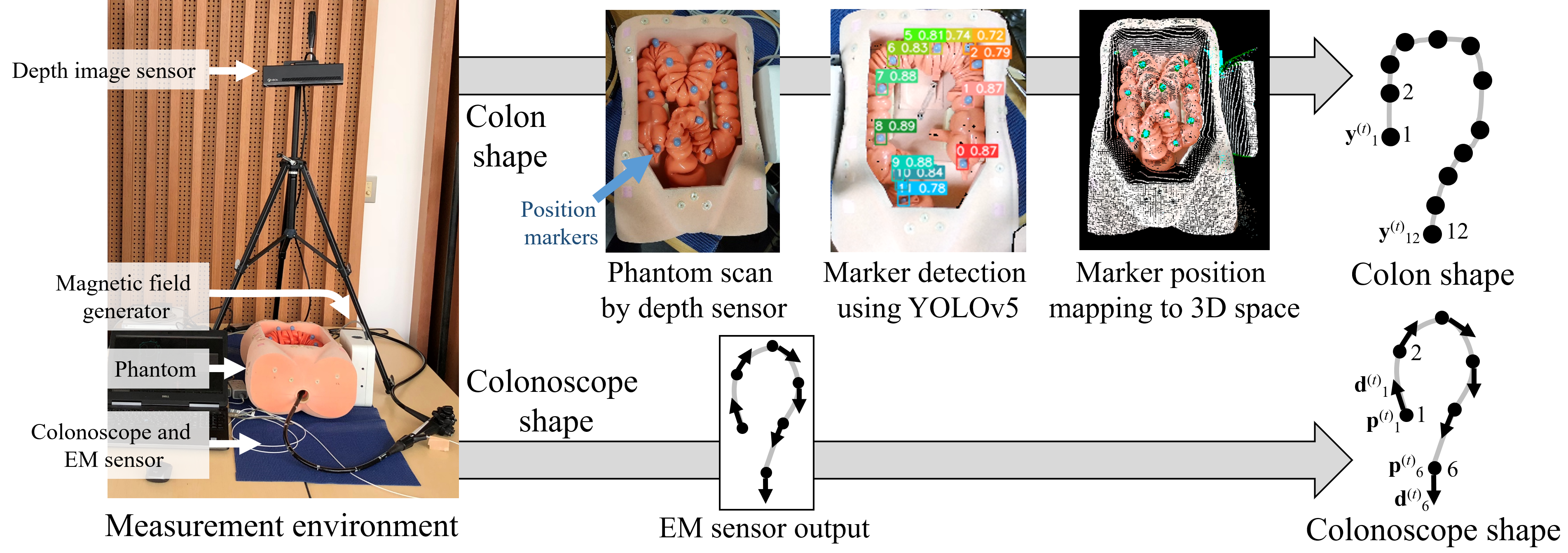}
\caption{Colonoscope and colon shapes measurement environment. Processing flow of measured data is also illustrated.} \label{fig:measurement}
\end{center}
\end{figure}

We evaluated the colon shape estimation accuracy of the proposed method in a phantom study.
We used a colon phantom (colonoscopy training model type I-B, Koken, Tokyo, Japan), a CT volume of the phantom, a colonoscope (CF-Q260AI, Olympus, Tokyo, Japan), an EM sensor (Aurora 5/6 DOF Shape Tool Type 1, NDI, Ontario, Canada), and a depth image sensor (Kinect v2, Microsoft, WA, USA).
We measured colonoscope and colon shapes in our measurement environment shown in Fig. \ref{fig:measurement}.
We assume the colonoscope tip is inserted up to the cecum when colonoscope tracking starts because physicians observe and treat the colon while retracting the colonoscope after its insertion up to the cecum.
The colonoscope was moved from the cecum to the anus.

\subsection{Colonoscope shape measurement}
We measured colonoscope shapes using the EM sensor.
The EM sensor is strap-shaped with six sensors at its tip and points along its strap-shaped body (one sensor is 6 DOF and remaining are 5 DOF).
Each sensor provides a 3D position and a 3D/2D direction along the colonoscope by inserting the sensor into the colonoscope working channel.
The measured colonoscope shape is ${\bf X}^{(t)} = \{ {\bf p}^{(t)}_{n}, {\bf d}^{(t)}_{n}; n=1,\ldots,6 \}$ at every time $t$.

\subsection{Colon shape measurement}
We measured colon shapes from the colon phantom using the depth image sensor.
Twelve position markers were attached to the surface of the colon phantom to scan its shape.
The depth image sensor scanned colon shapes during the colonoscope insertions to the colon phantom.
We automatically detect marker positions using YOLOv5 \citep{yolov5} from the scanned color and depth images.
Then, the detection results were manually corrected.
The detected markers are described as ${\bf Y}^{(t)}=\{ {\bf y}^{(t)}_{m}; m=1,\ldots,12 \}$ at every time $t$, used as colon shape.
${\bf y}^{(t)}_{1}$ and ${\bf y}^{(t)}_{12}$ correspond to the cecum and the anus positions, respectively.

\subsection{Training and testing of KST-Mixer}
We measured both ${\bf X}^{(t)}$ and ${\bf Y}^{(t)}$ during colonoscope insertions to the colon phantom.
The shape recording frequency was six times per second.
${\bf X}^{(t)}$ and ${\bf Y}^{(t)}$ belong to the EM and depth image sensor coordinate systems.
We registered them in the CT coordinate system using the iterative closest point (ICP) algorithm \citep{icp} and manual registrations.
Registered shape data was used to train and test the KST-Mixer.
Parameters used in the trainings were: $\tau=18$, $(s_{1},s_{2})=(6,3)$, $b=7$, $d_{1}=0.1$, $d_{2}=0.3$, $h_{S}=64$, $h_{C}=128$, 50 minibatch size, and 200 training epochs.
Mean squared error was used as the loss function in training.
We implemented the KST-Mixer using the Keras build in TensorFlow 2.4.0 running on a Windows PC equipped with a NVIDIA RTX A6000 GPU.
The KST-Mixer used 2.5 GBytes of GPU memory in trainings.

In the test step, we provide colonoscope shapes for testing to the trained KST-Mixer.
We obtain estimated colon shape ${\bf \hat{Y}}^{(t_{c})}$ of current time $t_{c}$ from it.

\section{Experimental Results}

We measured colonoscope and colon shapes during eight colonoscope insertions and recorded 1,388 shape pairs.
An engineering researcher operated the colonoscope.
Leave-one-colonoscope-insertion-out cross validation was performed for evaluation.
We used mean Euclidean distance (MED) (mm) between ${\bf Y}^{(t)}$ and ${\bf \hat{Y}}^{(t)}$ as an evaluation metric.
It is defined as
\begin{equation}
{\rm MED} = \frac{1}{12T} \sum^{T}_{t=1} \sum^{12}_{m=1} || {\bf \hat{y}}^{(t)}_{m} - {\bf y}^{(t)}_{m} ||.
\end{equation}

We compared the MED of the proposed method with previous colon shape estimation methods, including SEN (LSTM-based method) \citep{oda18} and regression forests-based method \citep{oda18-2}.
Table \ref{tab:evaluation} shows results of the comparison.
The proposed method achieved the smallest MED among the methods.
Statistical analysis of the results indicated that the proposed method significantly reduced MED compared to the SEN \citep{oda18} ($p<0.05$ with paired {\textit{t}-test of MED values).
We compared computation times in estimations of one colon shape among these methods.
The results are shown in Table \ref{tab:time}.
From the results, both the proposed and previous methods have real-time performances.
Examples of colon shape estimation results are in Fig. \ref{fig:results}.
The figure shows that the differences between the ground truths and estimated colon shapes were tiny.

\begin{table}[tb]
\begin{center}
\caption{Mean and standard deviation of MED calculated in cross-validations performed using proposed and previous methods.}\label{tab:evaluation}
\begin{tabular}{|c|c|}
\hline
Method & MED (Mean $\pm$ S.D.) (mm) \\
\hline
KST-Mixer (Proposed) & ${\bf 11.92 \pm 1.75}$ \\
SEN (LSTM-based method) \citep{oda18} & $12.58 \pm 2.08$ \\
Regression forests-based method \citep{oda18-2} & $13.08 \pm 1.55$ \\
\hline
\end{tabular}
\end{center}
\end{table}

\begin{table}[tb]
\begin{center}
\caption{Computation times of proposed and previous methods in estimation of one colon shape.}\label{tab:time}
\begin{tabular}{|c|c|}
\hline
Method & Computation time (msec.) \\
\hline
KST-Mixer (Proposed) & 7.3 \\
SEN (LSTM-based method) \citep{oda18} & 2.9 \\
Regression forests-based method \citep{oda18-2} & 8.9 \\
\hline
\end{tabular}
\end{center}
\end{table}

\begin{figure}[tb]
  \begin{minipage}{0.25\textwidth}
    \centering
\subfloat[]{\includegraphics[width=0.9\textwidth]{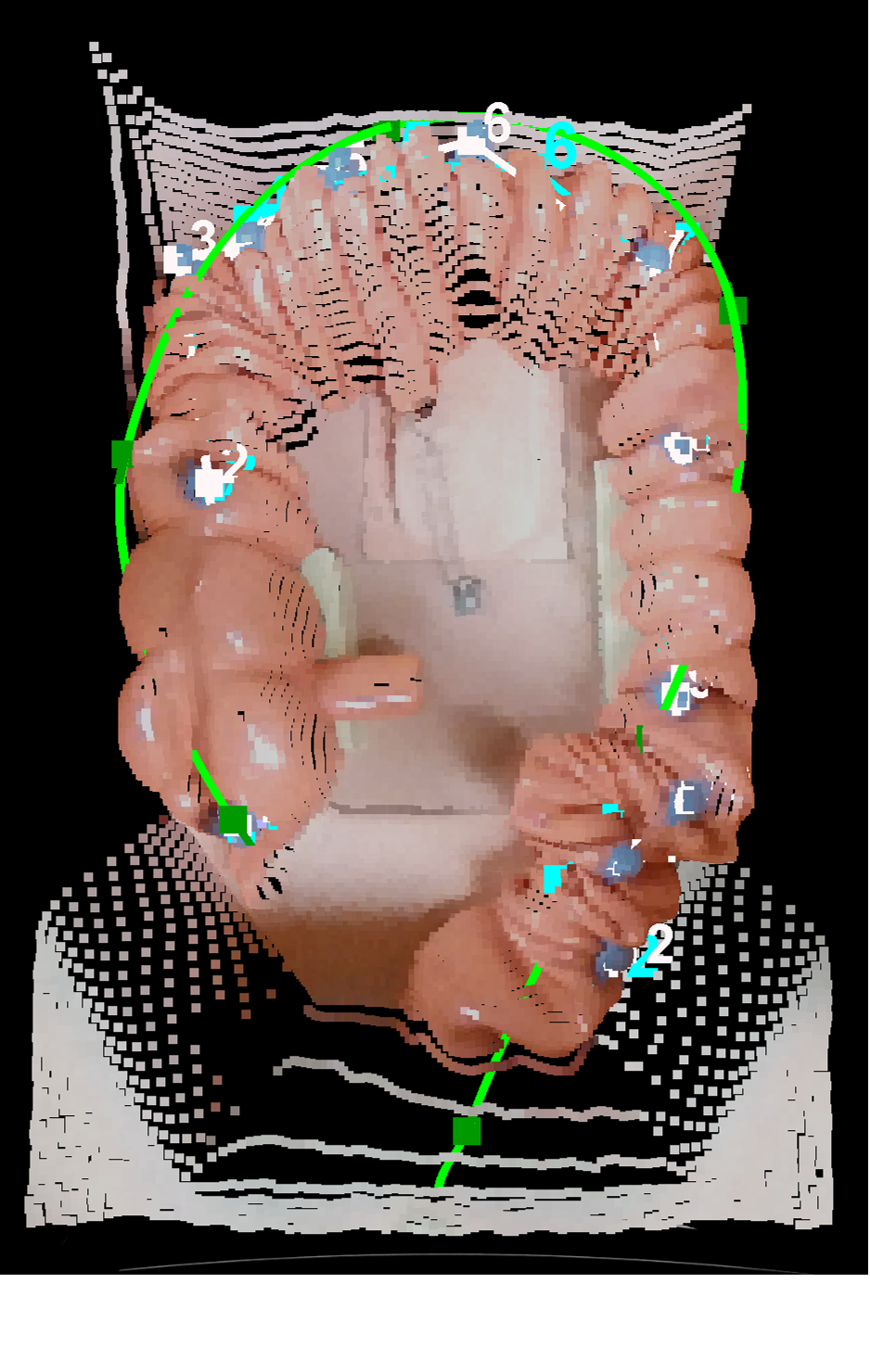}\label{fig:result_displayall}}
  \end{minipage}
  \begin{minipage}{0.75\textwidth}
\centering
\subfloat[]{\includegraphics[width=0.9\textwidth]{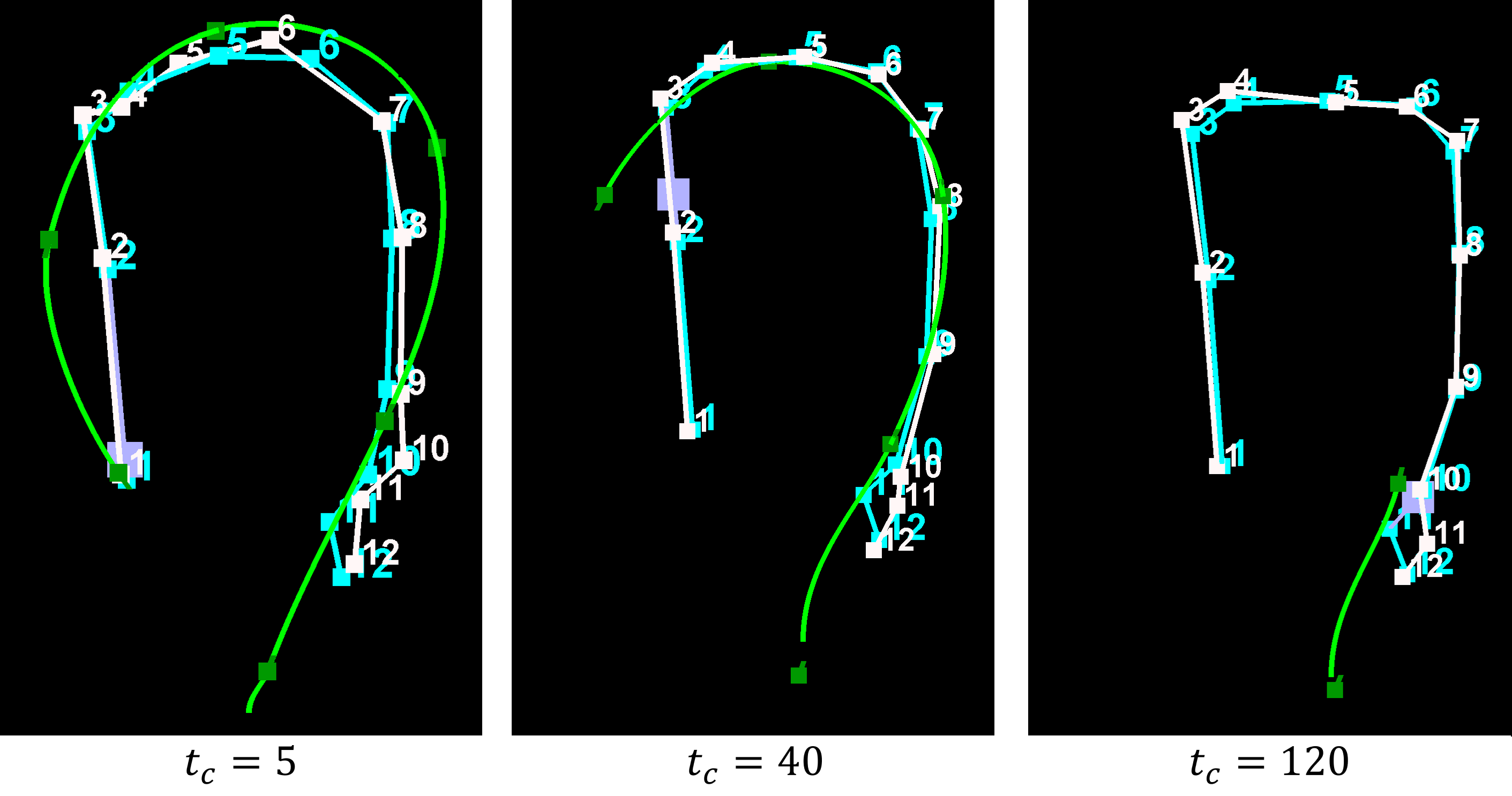}\label{fig:result_frames}}
  \end{minipage}
\caption{(a) Measurement results of depth image and EM sensors. (b) Colon shape estimation results of three frames. Colonoscope shapes are points on curved green lines. Colon shapes (ground truth) are white numbered points. Estimated colon shapes by proposed method are blue numbered points.} \label{fig:results}
\end{figure}

\section{Discussion}

The proposed KST-Mixer achieved the smallest error in the colon shape estimation among the previous methods.
We designed the KST-Mixer to extract features from kinematic data using simple MLP blocks.
Extracted features are then mixed in the spatio-temporal feature mixing MLP blocks to generate spatially and temporally global features.
This structure is quite effective in processing time-series kinematic data because we achieved better estimation results than the LSTM-based method \citep{oda18}.
The proposed method improves colonoscope tracking accuracy by accurately estimating deformed colon shape during colonoscope insertions.
Furthermore, the computation time of the proposed method was short enough to be used in real-time applications.

The application of the proposed method is not limited to colon shape estimation alone.
It can be applied to estimations of elastic organs in diagnosis and treatment.
Organ shape estimation is essential in surgical assistances by computers.
Accurate organ shape estimation contributes to the generation of real-time surgical navigation information and the automation of surgical assistance robots.

Although we have obtained promising results in colon shape estimation, many challenges are still remain for application of the proposed method to colonoscope tracking.
Such challenges include (1) {\it collecting data containing variations of operators and colon shapes}, (2) {\it collecting in-vivo data}, (3) {\it development of intuitive visualization method of tracking result}, and (4) {\it development of a colonoscope that have embedded EM sensors}.
(1) {\it collecting data containing variations of operators and colon shapes} is necessary to improve robustness of the method to real situations.
Colonoscope movements have variations among physicians depending on their years of experience.
Furthermore, colon shapes also have variations among patients.
Colon and colonoscope shapes data containing such variations is necessary to achieve better estimation model.
We will measure shape data under operations of colonoscope by physicians of various years of experience.
We also measure shape data using many colon phantoms and 3D printed phantoms that have variation of the shapes.
(2) {\it collecting in-vivo data} is necessary to improve the proposed method from phantom level to clinically applicable level.
(3) {\it development of intuitive visualization method of tracking result} is required that presents deformed colon shapes in real-time during colonoscope insertions.
Such visualization helps physicians to understand how the colonoscope traveling in the colon.
(4) {\it development of a colonoscope that have embedded EM sensors} is required to perform tracking in clinical situations.

\section{Conclusions}

This paper proposed a colon shape estimation method using the KST-Mixer from kinematic data.
The KST-Mixer extracts kinematic features and mixes them along the spatial and temporal axes in multiple MLP blocks.
We evaluated the method's estimation accuracy in the colon shape estimation from colonoscope shapes in the phantom study.
The proposed KST-Mixer achieved the smallest estimation error in the comparative experiments.
Future work includes improvement of the data number using other phantoms, evaluating the method using shape data measured during colonoscope operations by physicians, application to colonoscope tracking, and application to the human colon.


\section*{Disclosure statement}

No potential conflict of interest was reported by the author(s).

\section*{Funding}

Parts of this research were supported by the MEXT/JSPS KAKENHI Grant Numbers 21K12723, 17H00867 and the JSPS Bilateral International Collaboration Grants.

%

\end{document}